\documentclass[
    sigconf = true,     % MK: Style for GECCO proceedings (two columns and such).
    review = false,      % MK: Shows line number.
    screen = true,      % MK: Highlights links.
    anonymous = false,   % MK: Anonymous author information. (Also removes the Acknowledgment section.)
]
{acmart}

\usepackage{booktabs} % For formal tables

\acmYear{2020}
\setcopyright{none}
\acmConference[GECCO '20]{Genetic and Evolutionary Computation Conference}{July 8--12, 2020}{Cancún, Mexico}
\acmDOI{10.1145/3377930.3390189}
\acmISBN{978-1-4503-7128-5/20/07}

\usepackage{amsmath}
\usepackage{amssymb}
\usepackage{xspace}
\usepackage{xcolor}
\usepackage{dsfont}
\usepackage{rotating}
\usepackage{amsthm}
\usepackage{amsmath}
\renewcommand{\epsilon}{\varepsilon}

\newcommand{\A}{\mathcal{A}}

\newcommand{\VBSs}{VBS$_{\text{static}}$\xspace}
\newcommand{\VBSd}{VBS$_{\text{dyn}}$\xspace}

\DeclareMathOperator*{\argmin}{arg\,min}
\DeclareMathOperator*{\ERT}{ERT}

\begin{document}
\title[Towards Dynamic Algorithm Selection]{Towards Dynamic Algorithm Selection for Numerical\\ Black-Box Optimization: Investigating BBOB as a Use Case}
%{Towards Dynamic Algorithm Selection for Numerical\\ Black-Box Optimization: Investigating BBOB as a Use Case}
% \subtitle{Subtitle}

%%% The submitted version for review should be ANONYMOUS
\author{Diederick Vermetten}
\affiliation{
  \institution{Leiden Institute for Advanced Computer Science}
  \city{Leiden}
  \country{The Netherlands}
}
% \email{d.l.vermetten@liacs.leidenuniv.nl}

\author{Hao Wang}
\affiliation{
  \institution{Sorbonne Universit\'e, CNRS, LIP6}
  \city{Paris}
  \country{France}
}
% \email{hao.wang@lip6.fr}

\author{Thomas B{\"a}ck}
\affiliation{
  \institution{Leiden Institute for Advanced Computer Science}
  \city{Leiden}
  \country{The Netherlands}
}
% \email{t.h.w.back@liacs.leidenuniv.nl}

\author{Carola Doerr}
\affiliation{
  \institution{Sorbonne Universit\'e, CNRS, LIP6}
  \city{Paris}
  \country{France}}
% \email{Carola.Doerr@lip6.fr}

\renewcommand{\shortauthors}{D. Vermetten, H. Wang, T. B{\"a}ck, and C. Doerr}

\begin{abstract}
% \carola{Here as well, I suggest we remove the e-mail addresses -- everyone can find these easily, so that I do not see any benefit of having them. We also need to add these CSS classifiers (just pick one that seems OKish, maybe the same as last year, or browse the list, or check what other relevant papers have used)}
One of the most challenging problems in evolutionary computation is to select from its family of diverse solvers one that performs well on a given problem. This algorithm selection problem is complicated by the fact that different phases of the optimization process require different search behavior. While this can partly be controlled by the algorithm itself, there exist large differences between algorithm performance. It can therefore be beneficial to swap the configuration or even the entire algorithm during the run. Long deemed impractical, recent advances in Machine Learning and in exploratory landscape analysis give hope that this dynamic algorithm configuration~(dynAC) can eventually be solved by automatically trained configuration schedules. 
With this work we aim at promoting research on dynAC, by introducing a simpler variant that focuses only on switching between different algorithms, not configurations. Using the rich data from the Black Box Optimization Benchmark~(BBOB) platform, we show that even single-switch dynamic Algorithm selection (dynAS) can potentially result in significant performance gains. We also discuss key challenges in dynAS, and argue that the BBOB-framework can become a useful tool in overcoming these. 

\end{abstract}


% \ccsdesc[100]{Computing methodologies~Online learning settings}

\maketitle

\section{Introduction}

It is well known that, when solving an optimization problem, different stages of the process require different search behavior. For example, while exploration is needed in the initial phases, the algorithm needs to eventually converge to a solution (exploitation). State-of-the-art optimization algorithms therefore often incorporate mechanisms to adjust their search behavior \emph{while optimizing}, by taking into account the information obtained during the run. These techniques are studied under many different umbrellas, such as \emph{parameter control}~\cite{EibenHM99}, \emph{meta-heuristics}~\cite{metaheuristics_survey}, adaptive operator selection~\cite{AOS}, or \emph{hyper-heuristics}~\cite{BurkeGHKOOQ13}. The probably best-known and most widely used techniques for achieving a dynamic search behavior are the \emph{one-fifth success rule}~\cite{Rechenberg73,Devroye72,SchumerS68} and the \emph{covariance adaptation technique} that the family of CMA-ES algorithms~\cite{hansen_adapting_1996,hansen2001self_adaptation_es} is build upon. 
While each of these two control mechanisms tackles the problem of balancing performance in different phases of the search in its own way, they are mostly working with a specific algorithm, aiming to tune its performance by changing internal parameters or algorithm modules. This inherently limits the potential of these methods, since different algorithms can have widely varying performances during different phases of the optimization process. By switching between these algorithms during the search, these differences could potentially be exploited to get even better performance. We coin the problem of choosing which algorithms to switch between, and under which circumstances, the \emph{Dynamic Algorithm Selection}~(dynAS) problem. 
%\died{Is the following sentence beneficial?} 
% While dynAS is a challenging problem which has not yet been fully explored in the context of numerical optimization, techniques from other fields, such as reinforcement learning, might prove to be useful tools for initial research. 

% In this work, we use this higher level of abstraction offered by dynAS and ask how much benefit we could possibly see from implementing it. While this dynAS problem is not yet very well understood, it has the potential to significantly benefit from recent advancements in fields like reinforcement learning~\cite{BiedenkappBHL19} and landscape analysis~\cite{mersmann2011exploratory}. Techniques from these field could potentially be used to decide at what point in the search to switch to which algorithm. 

Solving the dynAS problem would be an important milestone towards tackling the more general \emph{dynamic Algorithm Configuration (dynAC)} problem, which 
%-- in addition to switching between a discrete set of different algorithms -- 
also addresses the problem of selecting (and possibly adjusting) suitable algorithm configurations. Specifically, dynAS is limited to switching between algorithms from a discrete portfolio of pre-configured heuristics, whereas for dynAC, the algorithms come with (possibly several) parameters whose settings can have significant influence on the performance. %that can be tuned. %(To be clear, we should say that we consider as AS the selection from a discrete portfolio of algorithms, whereas in AC all or at least some algorithms also come with parameters that can be set by the user.) 

We do not solve dynAS here, but aim to show its potential for numerical optimization. We then aim to develop suitable environments to encourage and enable future research into achieving the identified potential of dynAS and, in the longer run, to extend this to the dynAC problem. As a first step, we need to identify a meaningful collection of algorithms and benchmark problems, which together cover the main characteristics and challenges of the dynAS problem, without imposing too many additional challenges. The Black-Box Optimization Benchmarking (BBOB) environment~\cite{hansen_coco:_2016} with its rich data sets available at~\cite{bbob-data} suggests itself as a natural starting point for such considerations, since the community has already acquired a quite solid understanding of the problems and solvers in this test-bed over the last decade.  

We perform a first assessment of the performance that one could expect to see when applying dynAS to the algorithms in the BBOB data sets, to understand whether the gains would justify further exploration of the dynAS paradigm on this test-bed. We find that -- even when restricting the dynAS problem further to allowing only a single switch between algorithms in the portfolio -- promising improvements over the best static solvers can be expected, in particular for the more complex problems (functions 19-24).

Our considerations are purely based on a theoretical investigation of the potential, which might be too optimistic for the single-switch dynAS case -- most importantly, because of the problem of \emph{warm-staring} the algorithms: since the heuristics are adaptive themselves, their states need to be initialized appropriately at the switch. This may be a difficult problem when changing between algorithms of very different structure. We do not consider, on the other hand, the possibility to switch more than once, so that our bounds may be too too pessimistic for the full dynAS setting, in which an arbitrary number of switches is allowed. 

Given the above limitations, we therefore also provide a critical assessment of our approach, and highlight ideas for addressing the main challenges in dynAS. 
%Most notably, we use our findings to reduce the size of the full portfolio of \carola{XXX} solvers from the BBOB data repository, and to identify interesting algorithm combinations, for which a detailed investigation of dynAS would be meaningful.  

% However, we will note that these theoretical improvements can be too optimistic, since the evolution of running time does not fully capture the behaviour of the algorithms. We investigate possible pitfalls of algorithm switching in detail, and aim to provide a starting point for practical implementation of dynAS. All in all we believe that the BBOB data provides an interesting test bed for developing dynAS approaches. 
%which [have this or that property, e.g., weak or no global structure (or whatever - you will know better than me)]. (add a few disclaimers, or say that those will be discussed later on in the paper, and say that an obvious next step would be to implement and test these algos, but that this requires an own project, since BBOB does not offer any algorithm repository, so that we cannot simply test the dynamic approaches as easily as was done in~\cite{research_project}). All in all we believe that the BBOB data provides an interesting test bed for developing dynAS approaches.}

\subsection{Related Work}

The idea that a dynamic configurations and/or selection of algorithms can be beneficial in the context of iterative optimization heuristics is almost as old as evolutionary computation itself, in particular in the context of solving numerical optimization problems, see~\cite{LoboLM07} for an entire book focusing mostly on  dynamic algorithm configuration techniques. However, as mentioned above, existing works almost exclusively focus on changing parameters of selected components of an otherwise stable algorithmic framework. This includes most works on hyper-heuristics~\cite{BurkeGHKOOQ13} and related concepts such as adaptive operator selection~\cite{AOS}, and parameter control~\cite{EibenHM99}. 

To the best of our knowledge, the full dynAC problem as described above was only recently formalized~\cite{BiedenkappBHL19}. Biedenkamp \emph{et al.} introduce dynAC as a Contextual Markov Decision Process (CMDP), where a policy can be learned to switch hyperparameters of a meta-algorithm, with some of these hyperparameters possibly encoding the choice between different algorithms.\footnote{Note here that there is a long-standing debate about the classification of algorithm configuration vs. algorithm selection. That is, while some consider a parametrized algorithm framework an algorithm with different configurations, others argue that each such configuration is an algorithm by itself. We omit this discussion here, and use the convention that an algorithm can have possibly different configurations. Note, though, that -- in the context of this work -- this only makes a difference in the terminology. All concepts and ideas can be equivalently described using the other, possibly mathematically more stringent, convention.} They also show that artificial CMDPs can be solved effectively by using reinforcement learning techniques, providing a promising direction for future research on dynAC.
%They propose a set of 5 artificial benchmark functions for dynAC and show that reinforcement learning techniques provide promising results. \carola{I would provide more details here, since this work is quite relevant. We should say which problems (numerical, dicrete, MIP) and which solvers are studied and whether it is rather DynAS or rather DynAC, etc }

In the context of evolutionary computation, the concept of switching between different algorithms during the optimization process was recently investigated in~\cite{van_rijn_ppns_2018_adpative}, by a similar theoretical assessment as in this work. The approach was then tested in~\cite{research_project}, where it was shown that the predicted gains can indeed materialize, with the caveat that one has to ensure a sufficiently accurate estimate for the median anytime performances of each algorithm. These two works, however, focus on a single family of numerical black-box optimization techniques, the modular CMA-ES framework suggested in~\cite{van_rijn_evolving_2016}. Here in this work, in contrast, we explicitly want to go one step further, and study combinations of heuristics that are potentially of very different structure, such as, for example combining a Differential Evolution (DE) algorithm for the global exploration with a CMA-ES for the final convergence. 

While the dynAC problem is solved by an unsupervised reinforcement learning approach in~\cite{BiedenkappBHL19}, we observe that dynAC in evolutionary computation is more frequently based on on supervised learning approaches, see~\cite{MalanM19,MunozS17footprint,jankovic2019adaptive} for examples. These techniques combine exploratory landscape analysis~\cite{mersmann2011exploratory} and/or fitness landscape analysis~\cite{Pitzer2012fitnesslandscape} with supervised learning techniques, such as random forests, support vector machines, etc. While still in its infancy, even in the static algorithm configuration case~\cite{munoz2015algorithm, kerschke2017automated,KerschkeT19,BelkhirDSS17}, these works may pave an interesting alternative to reinforcement learning, as they may more directly provide insight into (and make use of) the correlation between fitness landscapes and algorithms' performance.

\section{Preliminaries}%\died{This section should probably have a different name}

\subsection{Dynamic Algorithm Selection}

Classically, algorithm selection attempts to find the best algorithm $A$ from a portfolio $\A$ to solve a specific function $f$ from a set of functions $\mathcal{F}$. Specifically, this static version of algorithm selection can be defined as follows: 
\begin{definition}[Static Algorithm Selection]
Given an algorithm portfolio $\A$ and a function $f\in\mathcal{F}$, we aim to find:
$$\argmin_{A\in\A} \text{PERF}(A, f) \; ,$$
where PERF is a performance measure (which assigns lower values to better performing algorithms).
\end{definition}

To extend algorithm selection to the dynamical case, we need to define a function which switches between algorithms. We use techniques from~\cite{BiedenkappBHL19} to represent this as a policy function, and modify it as follows:
\begin{definition}[Dynamic Algorithm Selection (dynAS)]%\died{Should we use DynAS or dynAS?}
Given an algorithm portfolio $\A$, a $f\in\mathcal{F}$ and a state description $s_t\in\mathds{S}$ at time step $t$ of an algorithm run. We want to find a policy $\pi: \mathds{S} \xrightarrow{} \A $ which minimizes $\text{PERF}(A_{\pi}, f)$
\end{definition}
Note that this definition can be extended to dynamic algorithm configuration by changing the policy to be $\pi: \mathds{S} \xrightarrow{} (\A\times\Theta_A) $, 
% \tb{Why $\otimes$? I think it should be $\times$}\carola{either way is fine for me, I do not really distinguish between those, to be honest}
where $\Theta_A$ is the configuration space of algorithm $A$.

\subsection{The BBOB Benchmark} %\carola{Remember to capitalize section headings: BBOB Competition Data (also applies to subsections, pargraphs etc)}
The Black Box Optimization Benchmark (BBOB) is widely accepted as the go-to benchmarking framework within the field of optimization. While BBOB has grown a lot over the years, the functions within their noiseless suite have remained stable. This suite contains 24 noiseless optimization functions, each of which being theoretically defined for any number of dimensions. In practice however, the commonly used dimension set is $\mathcal{D}=\{2,3,5,10,20,40\}$. For each function, several transformation methods are defined, both for the variable as the objective spaces. These transformations are fixed, and different combinations lead to different versions of the function, called instances. Since these functions are defined mathematically, the optimal values are known in advance. Because of this, we can define target values we wish to reach in terms of closeness to this optimal value, instead of an abstract value. This gives the advantage of comparability between instances, which would not be possible when using raw target values. 

The 24 noiseless functions have been studied in detail, not just from a performance perspective. Especially within the landscape analysis community, a lot of analysis of the BBOB-functions has been performed, leading to a lot of useful insights about their properties. These properties are ideal to use when implementing dynAS in practice, as they are very influential on the local performance of algorithms. Generally, it is agreed that the 24 BBOB functions cover a broad range of potential challenges for different optimization algorithms~\cite{mersmann2011exploratory}, even though certain aspects, i.e., discontinuities or plateaus, are not very well represented~\cite{lacroix2019}. 

The popularity of BBOB means that many researchers have benchmarked their algorithms on the BBOB-functions. Most of these have then submitted versions of their algorithms to competitions or workshops organized by the BBOB-team. Between the first competition in 2009~\cite{hansen2010comparing} and the latest workshop in 2019, a total of 226 algorithms have been submitted and their data made available to the public~\cite{bbob-data}. Because of this large amount of available data, there are plenty of baselines to compare algorithms against and gain inspiration from.  These algorithms have often been well justified and rigorously tested. However, the implementations used are generally not freely available, and even if they are, they might be hard to combine into a single dynAS framework, since BBOB is available in many different languages. However, the majority of the algorithms is either directly available online or has been well-documented, making the challenge of implementing them doable. 

Additionally, the large amount of algorithms which have been run on BBOB provide a good way to select sets of algorithms from which to build initial dynAS portfolios. However, since the BBOB-repository is largely the result from running competitions, many of the used algorithms are highly tuned, making them hard to beat and giving rise to the question of generalizability of dynAS results to other functions. Eventually, a move to true dynAC would resolve this issue, but these techniques will require a lot of further study to implement. 

Since the BBOB-framework provides the functions, algorithms and performance baselines, it is an ideal candidate for initial experiments related to dynAS. 

%An overview of these properties has been given in~\cite{mersmann2011exploratory} and is shown in Table~\ref{tab:bbobfucntions}. \carola{I think we do not need the table, but I do not have a strong opinion yet}

% \input{table_bbob.tex}

\subsection{Performance Measures}
To measure the performance of the algorithms on the BBOB-dataset, several approaches are possible. These usually fall into two categories: fixed-budget and fixed-target. The fixed-budget approach asks the question: "What target value is reached after $x$ function evaluations?", while the fixed-target question can be phrased as: "How many function evaluations are needed to reach target $y$?". 

In this paper, we will use the fixed-target approach. Since most algorithms in our data set are stochastic in nature, the question of how many function evaluations are needed to reach a certain target is dealing with random variables. For a certain function instance $f_i\in\mathcal{F}$ and dimension $d\in\mathcal{D}$, we let 
$t_j(A, f_i^{(d)}, \phi)$ 
denote the number of evaluations that algorithm $A\in\mathcal{A}$ needed in the $j$-th run to evaluate for the first time a point of target precision at least $\phi$. Note  that $t_j(A, f_i^{(d)}, \phi)$ is a random variable, which is commonly referred to as the \emph{Hitting Time (HT)}. If run $j$ did not manage to hit target $\phi$ within its allocated budget, we say that $t_j(A, f_i^{(d)}, \phi) = \infty$.
% \tb{$t_j(A, f_i^{(d)}, \phi) = \infty$}.

While just taking the average of the observed hitting time gives some estimate of the true mean, previous work~\cite{auger_restart_2005} has shown that it is not a consistent, unbiased estimator of the mean of the distribution of hitting times. Instead, the Expected Running Time (ERT) is used. This is defined as follows:
\begin{definition}[Expected Running Time (ERT)]
$$\ERT(A, f^{(d)}, \phi) = \frac{\sum_{i=1}^n\sum_{j=1}^K \min\{ t_{i}(A, f_j^{(d)}, \phi), B\}}{\sum_{i=1}^n\sum_{j=1}^K \mathds{1}\{t_{i}(A, f_j^{(d)}, \phi) <\infty\}}.$$
Here, $n$ is the number of runs of the algorithm, $K$ the number of instances of function $f$ and $B$ the maximum budget for algorithm $A$ on function $f_j^{(d)}$.
\end{definition}

To allow for a fair comparison between instances, the BBOB-benchmark uses target 'precisions' for their analysis, instead of the raw target values seen by the algorithm. The precision is simply defined as the difference between the best-so-far-$f(x)$ and the global optimum. This is done to make runtime comparisons between different instances and even different functions possible. 

\section{Methods}

\subsection{Analysis of Available data}\label{sec:preselection}
Since the set of available algorithms from the BBOB-competitions is quite large, several issues in terms of data consistency arise. When processing the algorithms, we found that a small subset have issues such as incomplete files or missing data. We decided to ignore these algorithms, and work only with the ones which were made available within the IOHanalyzer tool~\cite{iohprofiler}. This leaves us with a set of 182 out of 226 possible algorithms to do our analysis.

There are some caveats to this data, mostly related to the lack of a consistent policy for submission to the competitions over the years. For example, the 2009 competition required submission of 3 runs on 5 instances each, while the 2010 version changed this to 1 run on 15 instances. In theory, the instances should have very little impact on the performance of the algorithms, as they are selected in such a way to preserve the characteristics of the functions. However, in practice there has been some debate about the impact of instances on algorithm performance, claiming that the landscapes of different instances of the same function can look significantly different to an algorithm~\cite{Munoz18instances,MunozKH15InformationContent,KerschkeT19}. 
% \tb{do we have a reference?} \carola{I have added three which explicitly discuss differences between the instances, but I do not know from the top of my head where differences in performance have been reported}. \died{I know there is a figure in my thesis which shows some performance differences between instances, would that be sufficient?} \carola{Hm, I would prefer if we find another one, as this one may reveal our identity (and it is also only in the appendix of your thesis, if I remember correctly. If we do not find one (it has low priority), we can stick to the papers that I have cited} 
%For the purpose of this paper however, we will 
In the following, we ignore this discussion and assume that performance is not significantly impacted by the instances.

Another issue with the dataset are the widely inconsistent budgets for the different algorithms. These can be as low as $50D$ and as large as $10^7D$. However, since we use a fixed-target perspective to study the performance of the algorithms, these differences are not very impactful.

\begin{figure}
    \centering
    \includegraphics[width=0.5\textwidth, trim={0, 5, 0, 3},clip]{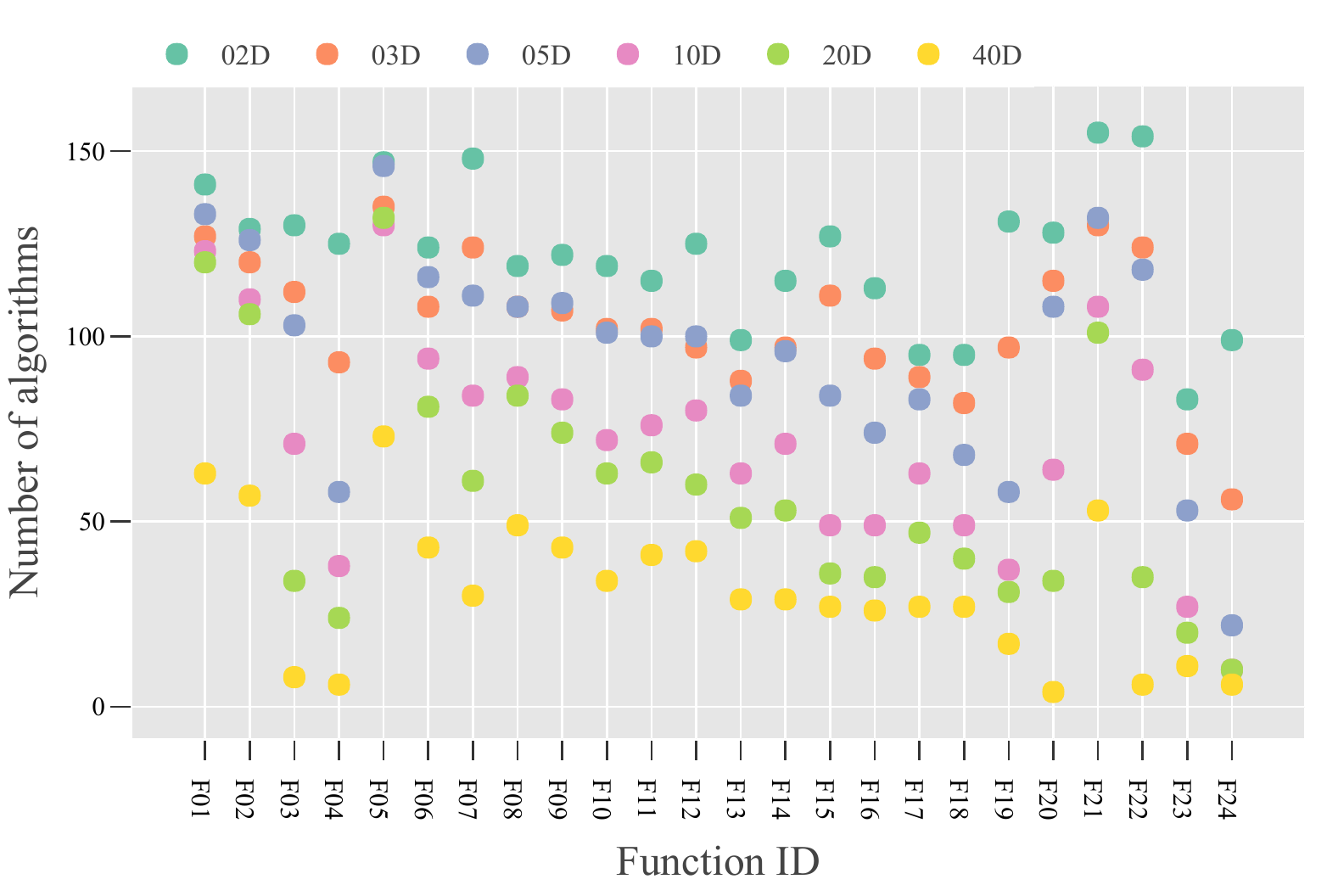}
    \caption{Number of algorithms with at least 15 independent runs and at least one them reaching the target $\phi=10^{-8}$.}\vspace{-10pt} %Distribution of algorithm which reached target $\phi=10^{-8}$ for each function and each dimension.\died{This figure should probably be plotted differently}
    % \carola{I would move the legend to the top, increase its font size, remove the D and write dimension somewhere. I like this figure very much and do not think the plotting needs to change!}
    \label{fig:nr_finished_algs}
\end{figure}

Since the BBOB-competitions see an optimizer as having 'solved' an optimization problem when reaching a target precision of $10^{-8}$, many of the algorithms will stop their runs after reaching this point to avoid unnecessary computation. Because of this, we will use the same target value in our computations. However, for some of the more difficult functions, this target can be challenging to reach within their budget. To avoid the problem of dealing with algorithms without any finished runs, we only consider an algorithm in our analysis when it has at least 15 runs on the function, of which at least one managed to reach the target $10^{-8}$. %The amount of algorithms this leaves us with for each function/dimension pair is plotted in 
Figure~\ref{fig:nr_finished_algs} plots the number of algorithms per each function/dimension pair that satisfy all the requirements mentioned above. We observe large discrepancies between functions and dimensions, with the number of admissible algorithms ranging from 4 to 155, and note that there are no algorithms which are admissible on all functions in all dimensions.
%an amount of available algorithms being anywhere between $4$ and $155$.

\subsection{DynAS for BBOB-Functions}\label{sec:dynas_bbob}

In this work, we will restrict the dynAS problem on BBOB-functions to using policies which switch algorithms based on the target precisions hit. To get an indication for the amount of improvement which can be gained by dynAC over static algorithm configuration,
%\tb{Where has this abbreviation been introduced?}\carola{Oh,since those are  important we should make sure, Diederick, that they are well visible :-) }
we use the BBOB-data to theoretically simulate a simple policy which only implements a single switch of algorithm. We can define this as follows:
\begin{definition}[Single-Switch dynAS]
Let $f^{(d)}$ be a BBOB-function in dimension $d$ and $\A$ the corresponding portfolio of admissible algorithms. A single-split policy is defined as the triple $(A_1, A_2, \tau) \in \A\times\A\times\Phi$, %\carola{$\in \A\otimes\A\otimes\Phi$, i.e. no brackets here (and in the following)}
where $\Phi = \left\{10^{2-0.2i)} | i\in \{0,\dots, 50\}\right\}$ is the set of admissible splitpoints. This corresponds to the policy which starts the optimization procedure with algorithm $A_1$, and run this until target $\tau$ is reached, after which the algorithm is changed to $A_2$.
\end{definition}

The performance of this single switch method can then be calculated as follows:
\begin{align*}
    T(f^{(d)},A_1,A_2, \tau, \phi) &= \ERT(A_1, f^{(d)}, \tau) \\ &+ \ERT(A_2, f^{(d)},\phi) - \ERT(A_2,f^{(d)},\tau)
\end{align*}
Where $\phi$ is the final target precision we want to reach. For the BBOB-functions, we set $\phi = 10^{-8}$, as noted in Section~\ref{sec:preselection}. 
% Note that $10^{-8}\in\Phi$, so for $A_1\in\A$ and $\tau = 10^{-8}$, we get for all $A_2\in\A$: $$T(f^{(d)},A_1,A_2, \tau) = \ERT(f^{(d)}, A_1, 10^{-8})$$ 
% \tb{I would write in the lhs $10^{-8}$ as well, not $\tau$.}\carola{I would delete that last sentence, I do not think it is particularly interesting, or am I missing something? Also, wouldn't it be better to define the performance more generally as $T(f^{(d)},A_1,A_2, \tau,\phi)$, to make the dependence on $\phi$ visible? And than we can say afterwards that here, according to the discussion above, we have fixed $\phi=10^{-8}$ and hence omit explicit mention of the final target precision.}

% \subsection{Static and Dynamic Best Solvers}

Generally, to assess the performance of an \textit{algorithm selection} method, its performance can be compared to the \textit{Single Best Solver (SBS)}, which can be defined as follows:
\begin{definition}[Single Best Solver]
For each dimension $d\in\mathcal{D}$, we have:
$$\text{SBS}_{\text{static}} (\mathcal{F}^{(d)})=\arg\min_{A\in\mathcal{A}}\sum_{f\in\mathcal{F}} \text{PERF}(A, f^{(d)}, \phi)$$
% Where PERF is a performance function.
% \tb{This has been said already (PERF); drop sentence and then write "Often, PERF = ERT, ...". However, previously PERF has been introduced with two arguments only! So:
% $\text{PERF}(A, f^{(d)}) =  \text{ERT}(f^{(d)}, A, 10^{-8})$
% Also, we might consider to make def. 3.2 more generic by replacing $10^{-8}$ by $\tau$.}.
Often, ERT is used as the performance function, but this value can differ widely between functions, leading to a biased weighting. To avoid this, we can instead use the ranking of ERT per function, to give equal importance to every function. Note that we have final target precision $\phi=10^{-8}$.
% \tb{$\tau$, not $\phi$.} \carola{Diederick, I suggest we always speak of ``final target precision $\phi$'' and switching point $\tau$ -- it is anyway good practice to always recall what a variable represents. Also helps readers to remember them.}
% as the target to reach, as is convention for the BBOB-framework, which will be further elaborated on in Section~\ref{sec:preselection}.
\end{definition}

While this SBS has a good average performance, it can easily be beaten by a decent \textit{algorithm selection} technique. As such, a better baseline for performance is needed. This is the theoretically best algorithm selection method, which is called the Virtual Best Solver. This can defined as follows:
\begin{definition}[Static Virtual Best Solver (\VBSs)]
For each function $f\in\mathcal{F}$ and dimension $d\in\mathcal{D}$, we have:
$$\text{VBS}_{\text{static}} (f^{(d)})=\arg\min_{A\in\mathcal{A}}\text{PERF}(A, f^{(d)})$$
For the BBOB functions, we use $\text{PERF}(A, f^{(d)}) = \ERT(A, f^{(d)}, \phi)$ with $\phi = 10^{-8}$.
\end{definition}
% \tb{So, here we use ERT right away instead of PERF!}

\begin{figure}
    \centering
    \includegraphics[width=0.5\textwidth, trim={0, 15, 0, 10},clip]{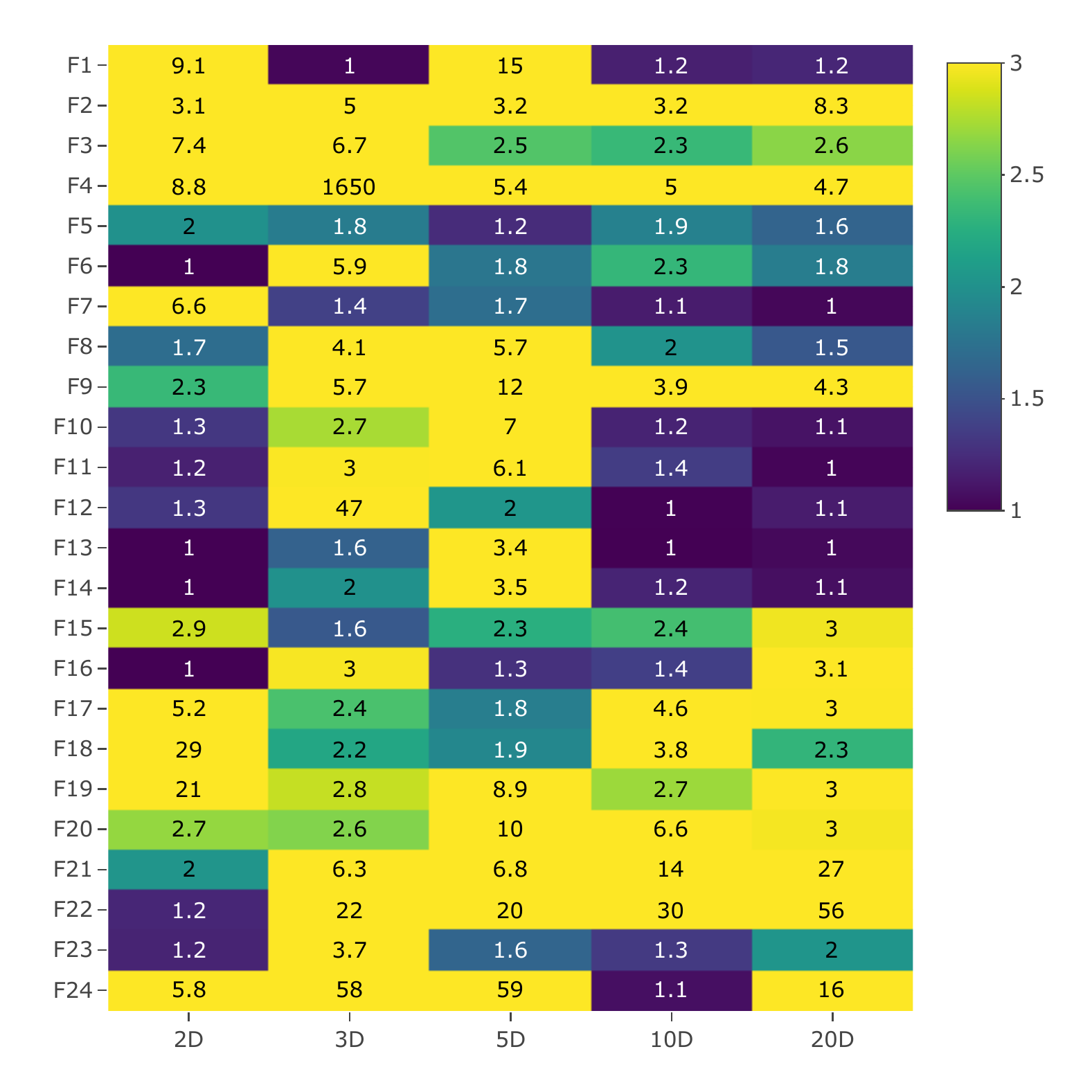}
    \caption{Relative ERT of the SBS over the \VBSs. The selected SBS are: Nelder-Doerr (2D), HCMA(3, 10 and 20D) and BIPOP-aCMA-STEP (5D). Dimension 40 was removed because no algorithm hit the final target on all functions in this dimension.}%, as we describe in Section~\ref{sec:preselection} \carola{I am confused -- shouldn't ERT of VBS be better? You probably plot ERT of SBS over ERT of VBS, no?}}
    \label{fig:sbs_vs_vbs}\vspace{-5pt}
\end{figure}

Note that the \VBSs will always perform at least as good as the SBS, and theoretically gives an upper bound for the performance of any real implementation of algorithm selection techniques. Thus, the difference between SBS and \VBSs gives an indication of the maximal possible performance gained by algorithm selection. For the BBOB-data, the relative ERT between these two methods is visualized in Figure~\ref{fig:sbs_vs_vbs}. From this, we see that the differences can be extremely large, highlighting the importance of algorithm selection. 

Similar to the way we defined \VBSs, we can define a Dynamic Virtual Best Solver, \VBSd, as follows:
\begin{definition}[Dynamic Virtual Best Solver]
For each BBOB-function $f\in\mathcal{F}$ and dimension $d\in\mathcal{D}$, we have:
$$\text{VBS}_{\text{dyn}} (f^{(d)})=\argmin_{(A_1,A_2, \tau)\in(\mathcal{A}\times\mathcal{A}\times\Phi)} T(f^{(d)}, A_1, A_2, \tau, \phi) $$
% \tb{Later on, we use dyn, not dynamic as the subscript to VBS.}
% Where $\Phi = \left\{10^{2-(0.2\cdot i)} | i\in \{0,\dots, 50\}\right\}$ is the set of possible splitpoints. 
\end{definition}
% \died{TODO: glue this section together in a better way}

\begin{table*}[!bht]
\tiny
\centering
\begin{tabular}{rlrllrrr}
  \hline
FID & \VBSs & ERT of \VBSs & $A_1$ & $A_2$ & $\log_{10}(\tau)$ & ERT of \VBSd & speedup \\ 
  \hline
1 & fminunc & 13.0 & HMLSL & HCMA & 1.2 & 6.6 & 1.97 \\ 
  2 & LSfminbnd & 94.7 & BrentSTEPrr & LSfminbnd & 2.0 & 52.4 & 1.81 \\ 
  3 & BrentSTEPrr & 315.5 & STEPrr & BrentSTEPif & -0.2 & 246.8 & 1.28 \\ 
  4 & BrentSTEPif & 763.9 & STEPrr & BrentSTEPif & -0.2 & 578.1 & 1.32 \\ 
  5 & MCS & 10.8 & ALPS & MCS & 1.8 & 6.0 & 1.80 \\ 
  6 & MLSL & 1050.9 & fmincon & GLOBAL & -7.0 & 928.2 & 1.13 \\ 
  7 & PSA-CMA-ES & 1129.8 & GP5-CMAES & PSA-CMA-ES & 0.0 & 792.3 & 1.43 \\ 
  8 & fminunc & 399.1 & OQNLP & DE-BFGS & 0.6 & 304.7 & 1.31 \\ 
  9 & fminunc & 188.3 & fminunc & DE-AUTO & 0.0 & 152.3 & 1.24 \\ 
  10 & DTS-CMA-ES & 262.4 & fmincon & DTS-CMA-ES & -2.0 & 199.8 & 1.31 \\ 
  11 & DTS-CMA-ES & 268.3 & HMLSL & DTS-CMA-ES & -2.2 & 153.6 & 1.75 \\ 
  12 & NELDERDOERR & 1909.7 & HMLSL & BFGS-P-StPt & -3.2 & 1041.5 & 1.83 \\ 
  13 & IPOPsaACM & 835.1 & DE-AUTO & IPOPsaACM & -3.6 & 661.7 & 1.26 \\ 
  14 & DTS-CMA-ES & 546.6 & DE-BFGS & DE-SIMPLEX & -6.0 & 348.6 & 1.57 \\ 
  15 & PSA-CMA-ES & 10029.7 & LHD-10xDefault-MATSuMoTo & PSA-CMA-ES & 0.4 & 6982.4 & 1.44 \\ 
  16 & IPOPsaACM & 6767.1 & GLOBAL & CMA-ES-TPA & -0.4 & 5115.0 & 1.32 \\ 
  17 & PSA-CMA-ES & 4862.3 & PSA-CMA-ES & IPOP400D & -5.8 & 4201.8 & 1.16 \\ 
  18 & PSA-CMA-ES & 6717.4 & PSA-CMA-ES & CMA-ES multistart & -5.2 & 5687.3 & 1.18 \\ 
  19 & DTS-CMA-ES & 18768.0 & OQNLP & DTS-CMA-ES & -1.6 & 463.0 & 40.54 \\ 
  20 & DEctpb & 10670.3 & DEctpb & OQNLP & -0.4 & 3360.7 & 3.18 \\ 
  21 & GLOBAL & 2095.5 & MLSL & NELDERDOERR & 0.0 & 1209.8 & 1.73 \\ 
  22 & GLOBAL & 1079.9 & RAND-2xDefault-MATSuMoTo & GLOBAL & 0.4 & 844.1 & 1.28 \\ 
  23 & CMA-ES-MSR & 18971.4 & DTS-CMA-ES & SSEABC & -2.6 & 10295.0 & 1.84 \\ 
  24 & OQNLP & 285173.0 & GP5-CMAES & CMAES-APOP-Var2 & 0.0 & 52387.0 & 5.44 \\ 
   \hline
\end{tabular}
\caption{Relative gain of optimal single-switch dynamic algorithm combination \VBSd over the best static algorithm \VBSs for all 24 BBOB functions in dimension 5. ERT values are computed from data available at \url{https://coco.gforge.inria.fr/doku.php?id=algorithms-bbob}. We only consider algorithms with at least 15 runs, one of which reaching target precision $\phi=10^{-8}$, which is also the target used for the ERT calculations. The full version of this table, also for other dimensions, is available at~\cite{data-bbob-link}. Abbreviations: FID = function ID (as in~\cite{hansen_coco:_2016}, $\tau$ = splitpoint target, speedup = ERT\_{stat}/ERT\_{dyn}. We also shortened DTS-CMA-ES\_005-2pop\_v26\_1model to DTS-CMA-ES for readability} 
\label{Dim5_overview_table}\vspace{-20pt}
\end{table*}

\section{Results}
Since the number of algorithms considered in this paper is relatively large, many of the results are only shown for a subset of functions, dimensions or algorithms. The complete data is made available at~\cite{data-bbob-link}. An example of the available data is also shown in Table~\ref{Dim5_overview_table}.
%\carola{We can consider to provide (here or in the appendix) an example for how the data looks like, e.g., in a table with the following columns dim, fID, Algo A1, Algo A2, $\tau^*(A1,A2)$, ERT(A1,$\tau^*$), ERT(A2,$\tau^*$), ERT(A2,$10^{-8}$), ERT(A2,$10^{-8}$), ERT(A1,A2,$\tau^*$), ERT(SBS,dim,fID), ER(VBS,dim,fID), or whatever else you consider interesting. Just to give an impression of what the user can expect to find in the GitHub repository.}

% \carola{Parts of Sections 4.1 and 4.2 will be integrated into Section 2.4, whereas I would introduce a new Section 4.1 Overall Gain of 1-Switch DynAS. And then describe results from Fig 3 and 4.}

\subsection{Overall Gain of Single-Switch DynAS}
\begin{figure}
    \centering
    \includegraphics[width=0.48\textwidth, trim={0, 5, 0, 3},clip]{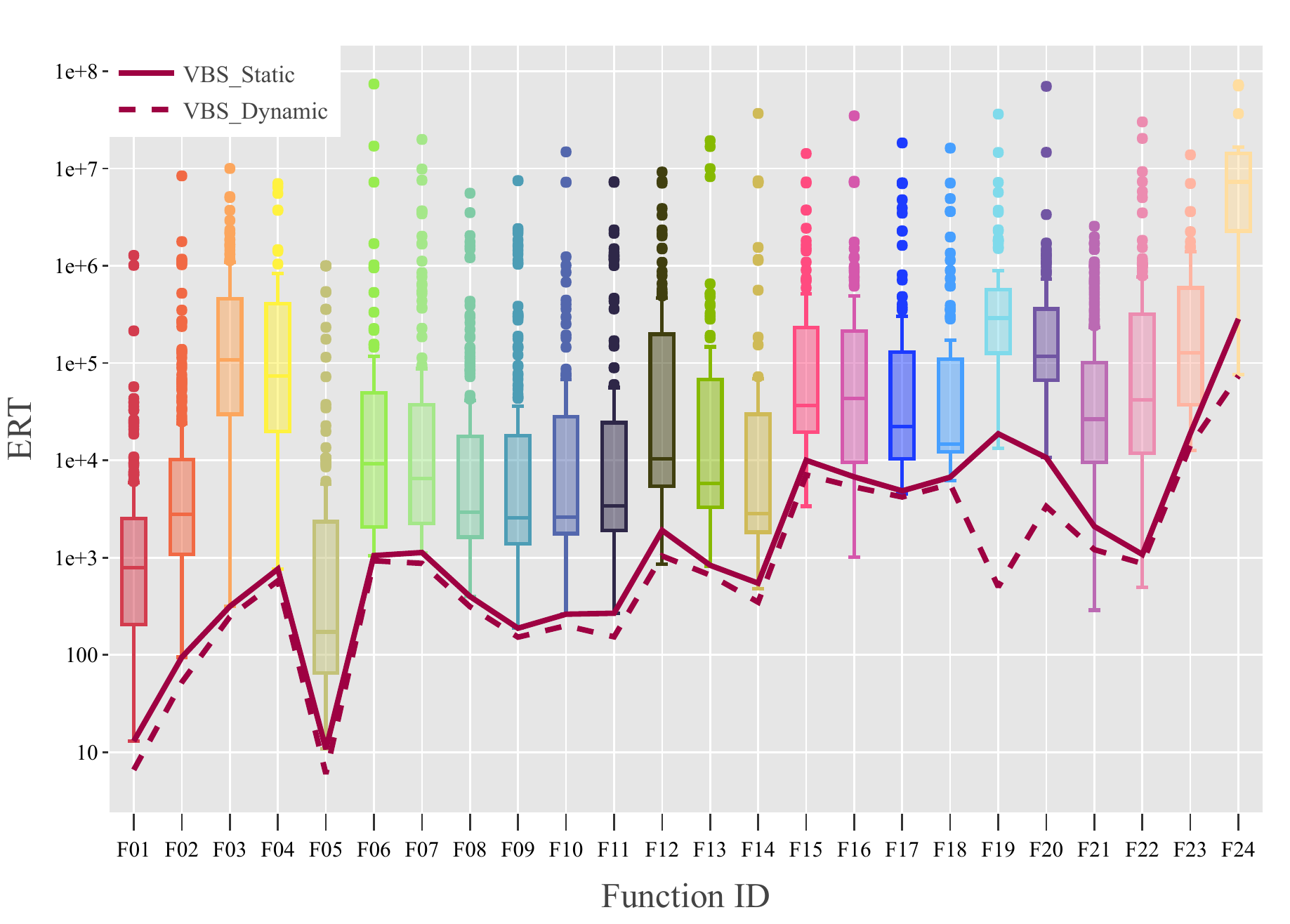}
    \caption{Distribution of ERTs among all algorithms for all 24 BBOB-functions in dimension 5. Please recall from Fig.~\ref{fig:nr_finished_algs} that the number of data points varies between functions. Also shown are the ERTs of the \VBSs and \VBSd.}%\carola{I like the data behind figure, but I would use classic box plots instead of these violins, which are somewhat strange, since they all show the same form, which I do not trust. Also, I suggest you add one line for the VBS.static and one for VBS.dyn }\died{I will update this to a version with the added lines later}}
    \label{fig:violin_static_ert_dim5}\vspace{-10pt}
\end{figure}

Before investigating the possible improvements to be gained by dynamic algorithm selection, we investigate the performance of the static algorithms from the BBOB-dataset. To achieve this, we look at the distribution of ERTs among the BBOB-functions. For dimension 5, this is visualized in Figure~\ref{fig:violin_static_ert_dim5}.\footnote{Note that for function F05, %\tb{should we write F05 to be consistent with the figures?}\carola{yes, agree}
the linear slope, most algorithms simply move outside the search-space to find an optimal solution, which is accepted by the BBOB-competitions, but leads to a disadvantage to those algorithms which respect the bounds.} This figure shows the large differences in performance, both between the algorithms as well as between the different functions. We marked the performance of the \VBSs and \VBSd, and see that their differences also vary largely between functions.  

% \begin{figure}
%     \centering
%     \includegraphics[width=0.5\textwidth]{Images/Violins_dim20.pdf}
%     \caption{Distribution of ERTs among all algorithms for all 24 BBOB-functions in dimension 20. Please recall from Fig.~\ref{fig:nr_finished_algs} that the number of data points varies between functions. Also shown are the ERTs of the \VBSs and \VBSd.}\died{Note that the whiskers currently are 1.5 times inter quartile range. The plotting software I use does not support changing this, so I will try to find some other way to fix this later.}
%     \label{fig:violin_static_ert_dim20}
% \end{figure}

% \subsection{Maximum Relative Improvement}

\begin{figure}
    \centering
    \includegraphics[width=0.45\textwidth, trim={0, 15, 0, 15},clip]{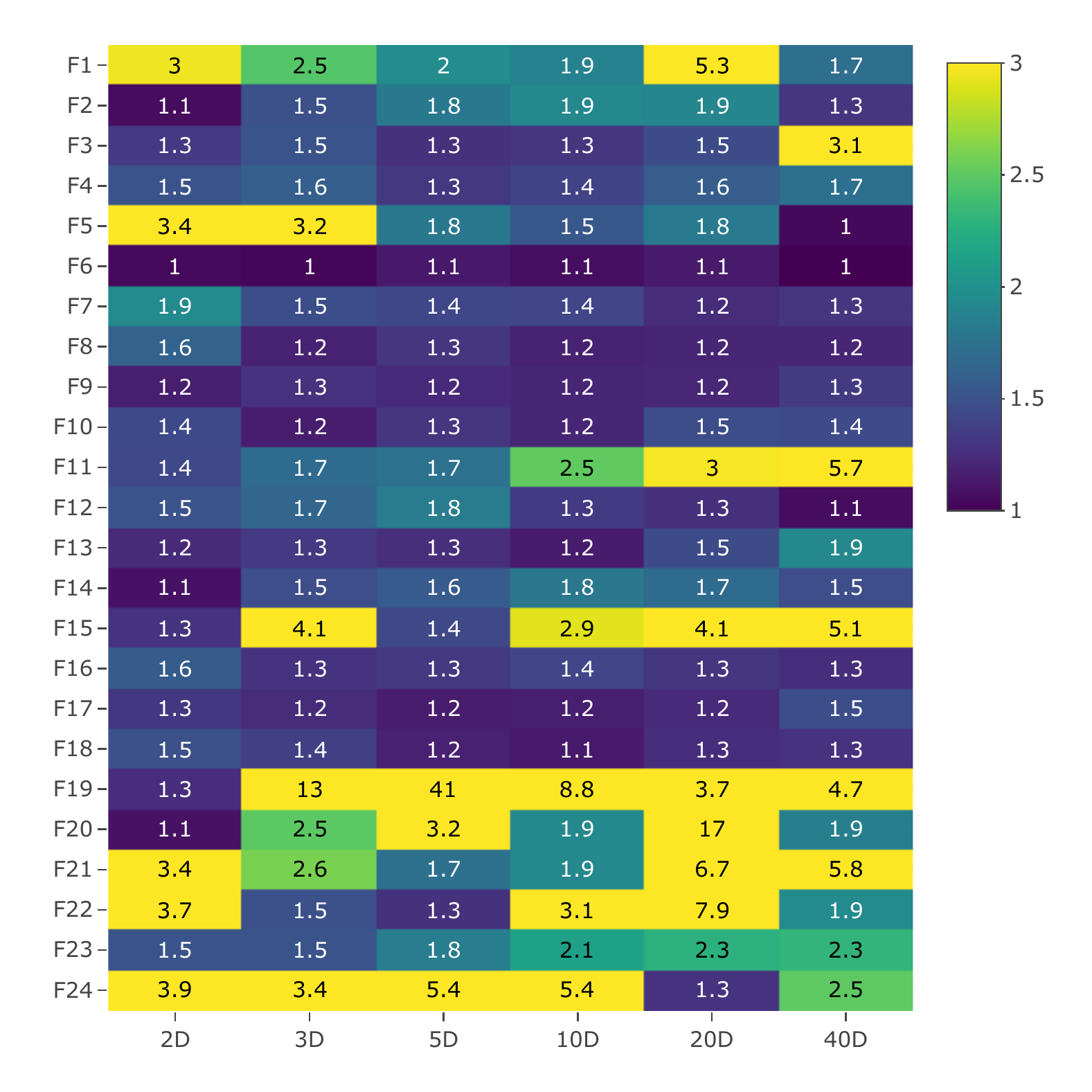}
    \caption{Heatmap of the ratio of ERTs between the Virtual Best Static Solver and the Virtual Best Dynamic Solver, for each (function, dimension)-pair. }%\carola{I would reverse the order of the rows (1 on top, 24 on bottom)}}
    \label{fig:heatmap_impr_dim_func}\vspace{-10pt}
\end{figure}

To zoom in on the differences between the \VBSs and \VBSd we see in Figure~\ref{fig:violin_static_ert_dim5}, we can compute for each function, dimension and corresponding algorithm portfolio the relative ERT of a the Single-Switch \VBSd over \VBSs. Specifically, this is calculated as $\frac{\ERT(\text{VBS}_{\text{dynamic}} (f^{(d)}))}{\ERT(\text{VBS}_{\text{static}} (f^{(d)}))}$. This value is shown for each (function, dimension)-pair in Figure~\ref{fig:heatmap_impr_dim_func}. From this figure, we can see that for most functions, the improvements when using a single configuration change are quite large. Especially for the functions which are traditionally considered more difficult for a black-box optimization algorithm to solve, the
possible improvement is massive. In terms of the median over all (function, dimension)-pairs, the \VBSd is $1.49$ faster than the \VBSs. 
% \tb{In terms of the median over all (function, dimension)-pairs, ...}
%Note: Mean is 2.47
% On average, the relative improvement of \VBSd over \VBSs is $38.9\%$. \died{TODO: Median of rel ERT} \died{Does this last sentence still belong here? Replacing it by the average relative ERT might be slightly misleading since the mean of relative ERTs is very prone to outliers (see SBS/VBS comparison)}

\subsection{Selected Algorithm Combinations} 

\begin{figure}
    \centering
    \includegraphics[width=0.47\textwidth, trim={0, 10, 0, 15},clip]{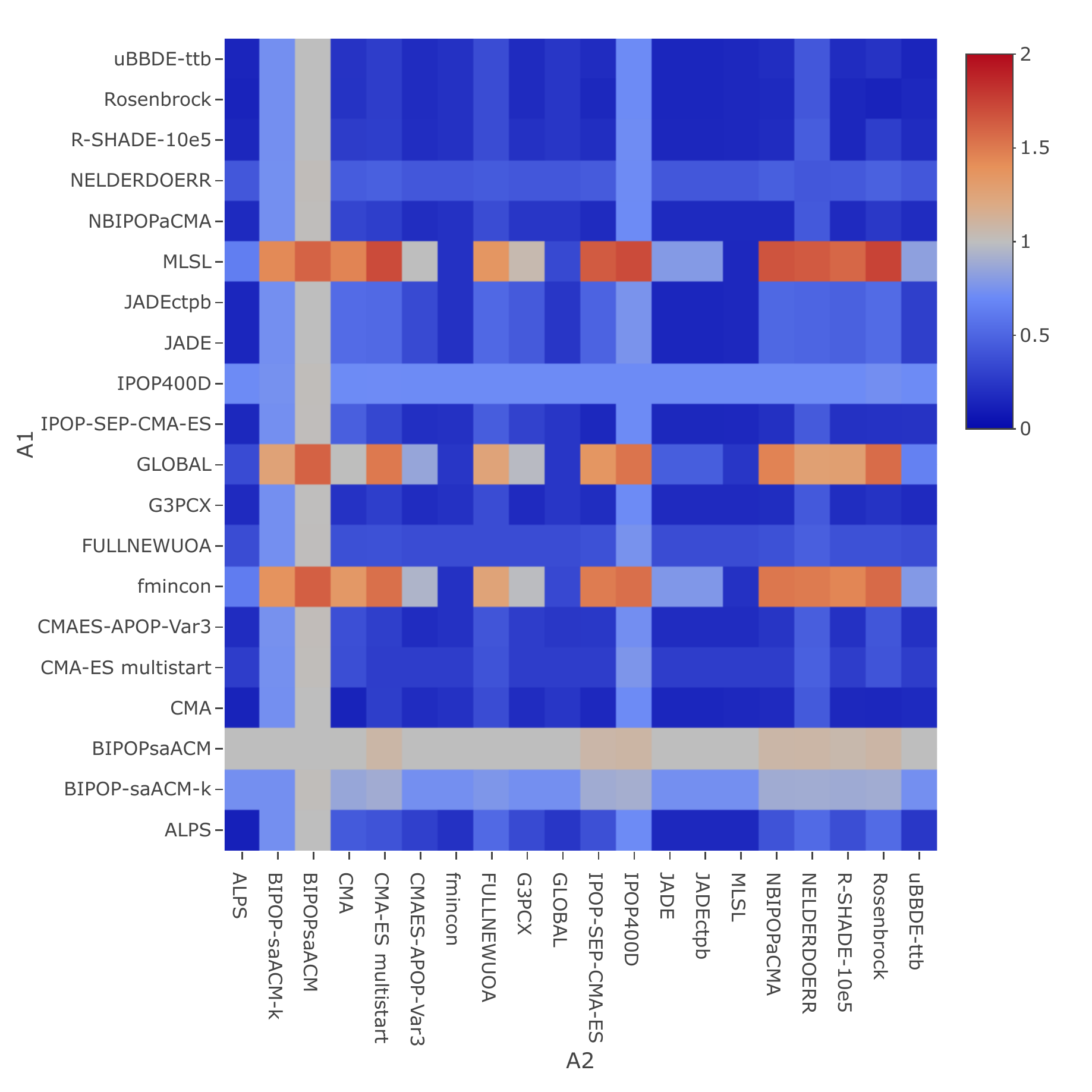}
    \caption{Relative ERT of configuration switches relative to \VBSs, for function 21 in 10 dimensions. The X- and Y-axes indicate algorithms selected as $A_2$ and $A_1$ respectively.  Larger values (red) indicate better algorithm combinations. }%\carola{I would swap that -- A1 should be y-axis, A2 on x-axis} \died{I indeed had the labels the wrong way around.}}%\carola{Is it correct that here we could use a color scheme that has a center at 1 and shows improvements (<1 values, correct?) in one color and values >1 in another one? With the current color scheme I cannot distinguish improvements over non-improvements }}
    \label{fig:F21_10_dim}\vspace{-10pt}
\end{figure}

% \carola{The first sentence needs to be revised, I do not quite get it} 
% While this initial data gives a good indication of the possibilities for dynamic algorithm configuration, the feasibility of the selected algorithm combinations might be challenging. 
Since the \VBSd shows a lot of potential improvement over the classical \VBSs, it makes sense to study its behaviour in more detail. To achieve this, we can zoom in on a single (function, dimension)-pair and study the behaviour of the \VBSd and split algorithm configurations in general. In Figure~\ref{fig:F21_10_dim}, we show the ERT of the best possible switch between any combination of algorithms in our portfolio $\mathcal{A}$, on function $21$ in dimension $10$. This figure shows some clear patterns in the horizontal and vertical lines. A horizontal line, such as the one for the MLSL-algorithm~\cite{MLSL1999}, indicates that an algorithm adds to the performance of most algorithms by being the $A_1$-algorithm. 
% \tb{Maybe I do not get this, but: blue is better, so maybe you mean vertical line (i.e., the column for MLSL indicates that it is a good $A_1$-algorithm since it has a lot of darker blue?} \died{I had the labels the wrong way around, but that is fixed now}
This can be interpreted as having a good exploratory search behaviour, but poor exploitation. There are also vertical lines present, which indicate the algorithms which perform well as $A_2$-algorithms. These are less pronounced than the horizontal lines, which might indicate that the choice of $A_2$ algorithms has less impact on the performance than the choice of $A_1$. %\died{Does this sentence make sense?}
%\died{TODO: choose a good example}.

% \begin{figure}
%     \centering
%     \includegraphics{}
%     \caption{Caption}
%     \label{fig:my_label}
% \end{figure}

We see that there are different algorithms which perform well as either the first or second part of the search. This gives rise to the question of how to quantify these differences, and more generally, how to quantify the benefit which can be gained by selecting an algorithm as $A_1$ or $A_2$. This can be done by executing the following steps to compute a quantitative value for the benefit gained by selecting an algorithm for a part of the search:
\begin{definition}[Improvement-values]
The initial performance value $I_1$ and finishing performance value $I_2$ of algorithm $A$ on function $f^{(d)}$ can be defined as:
$$I_1(A) = \frac{\min_{A_2\in\mathcal{A}, \tau \in \Phi} T(A,A_2,\tau, \phi)}{\min_{A_1,A_2\in\mathcal{A}, \tau \in \Phi} T(A_1,A_2,\tau, \phi)}$$
$$I_2(A) = \frac{\min_{A_1\in\mathcal{A}, \tau \in \Phi} T(A_1,A,\tau, \phi)}{\min_{A_1,A_2\in\mathcal{A}, \tau \in \Phi} T(A_1,A_2,\tau, \phi)}$$
\end{definition}
% \begin{itemize}
%     \item Let $f^{(d)}$ be the function to optimize. Take $A\in\A$ as the algorithm to use.
%     \item 
%     \item Set $I_1(A) = \frac{\min_{A_2\in\mathcal{A}, \tau in \Phi} T(A,A_2,\tau)}{\min_{A,A_2\in\mathcal{A}, \tau in \Phi} T(A,A_2,\tau)}$ %\carola{We should probably introduce the abbreviation for $T(f^{(d)},\mathcal{A},\Phi)={\min_{(A_1,A_2,\tau)\in\mathcal{A}\times \mathcal{A} \times \Phi} T(f,A_1,A_2,\tau)}$} \carola{Also, note here that I would not want to keep $\hat{\tau}$ fixed in the denominator.} \carola{One more thing, I think we should add $f$ to these definitions, i.e., $T(f,A_1,A_2,\tau)$ and not $T(A_1,A_2,\tau)$ (unless it becomes very inconvenient)} \carola{Last comment: maybe we should introduce all these terms or at least $T(f,A_1,A_2,\tau)$ before defining VBS.dyn, so that we can make use of that in the description of VBS (e.g., ``VBS is the algorithm that for each instance $f$ chooses a combination $(A_1,A_2,\tau)$ such that  $T(f,A_1,A_2,\tau) = T(f,\mathcal{A},\Phi)$) -- although that works only when $\mathcal{A}$ and $\Phi$ are finite, but since we are in this case, it's OK to use that definition}
% \end{itemize}
% We will refer to this value  $I_1(A)$ as the \died{TODO: Good name for this value} of algorithm $A$. We can define $I_2(A)$ analogously, by fixing $A$ as the algorithm to switch to, and looking at all possible $A_1$'s. %using the same procedure. 
Note that for the \VBSd$=(A_1,A_2,\tau)$, we always have $I_1(A_1)=1=I_2(A_2)$, and values can not be below $1$. Intuitively, the larger the value of $I_1$, the worse the algorithm can perform as the first part of the search, and similarly for $I_2$.

\begin{figure}
    \centering
    \includegraphics[width=0.5\textwidth, trim={0, 5, 0, 15},clip]{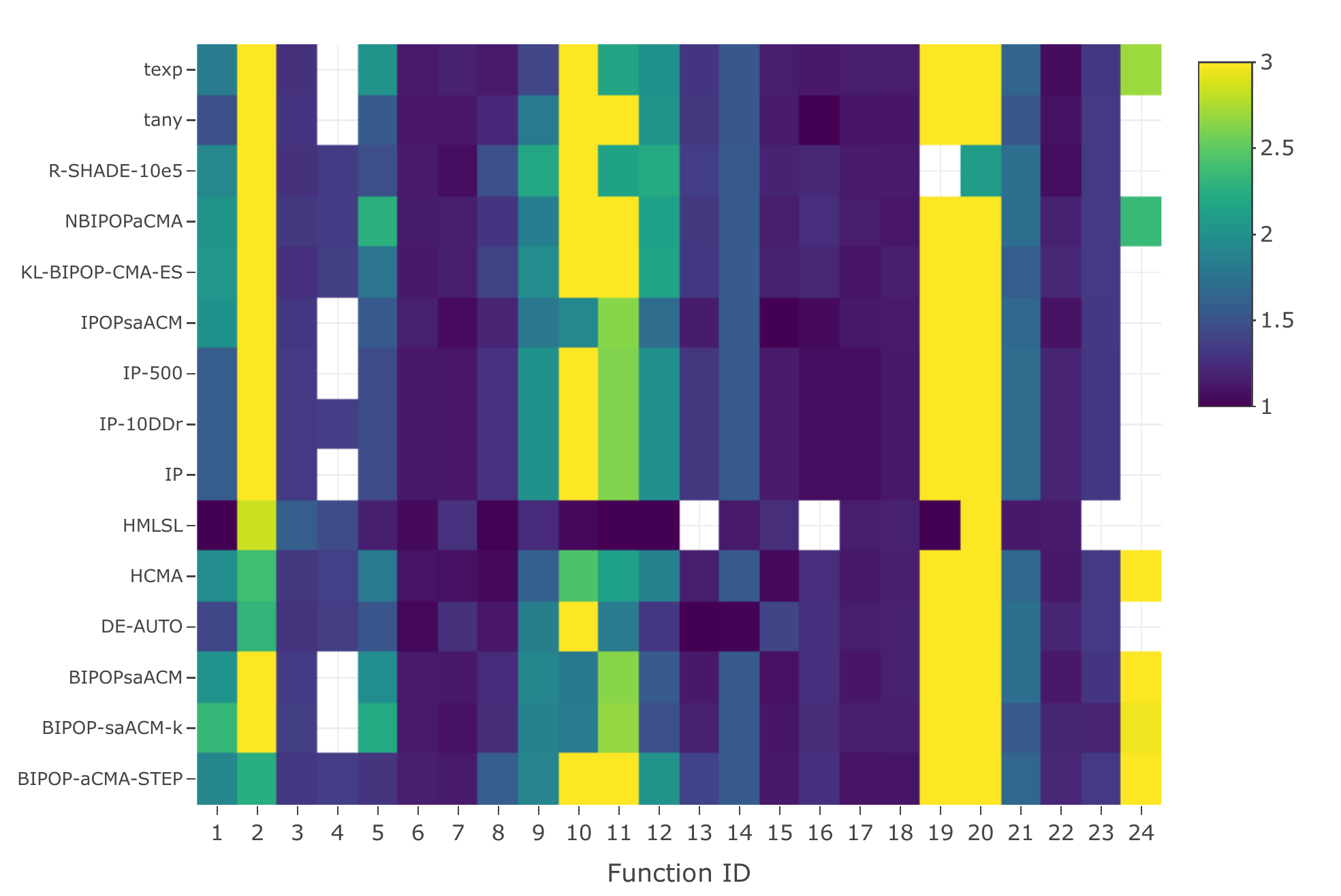}
    \caption{$I_1$-values for a group of 15 selected algorithms in dimension $5$. Darker colors correspond to better values.}
    \label{fig:I1_values_subset}\vspace{-5pt}
\end{figure}

\begin{figure}
    \centering
    \includegraphics[width=0.47\textwidth, trim={0, 5, 0, 15},clip]{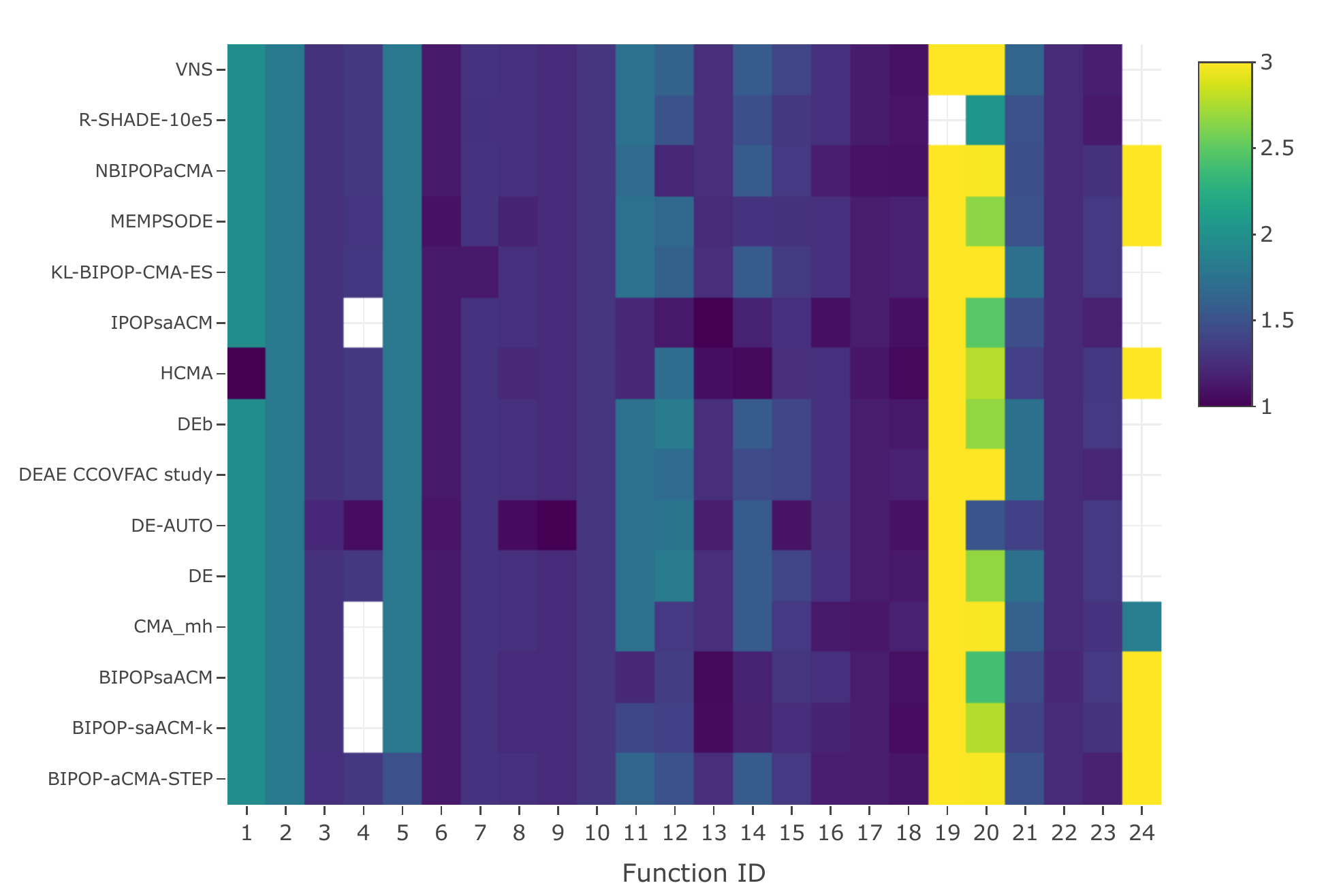}
    \caption{$I_2$-values for a group of 15 selected algorithms in dimension $5$. Darker colors correspond to better values.}% \carola{add axis label for x-axis: Function ID}}
    \label{fig:I2_values_subset}\vspace{-5pt}
\end{figure}

The values of $I_1$ and $I_2$ for dimension $5$ are shown in Figures~\ref{fig:I1_values_subset} and~\ref{fig:I2_values_subset} respectively. To ensure the readability of the figures, only a subset of algorithms is chosen. This is done by selecting the algorithm with the best value for each function, and then adding to it the set of algorithms which have the best average value over all functions\footnote{Missing values and values larger than $3$ are set to $3$ to reduce the large impact of outliers on the average.}.  From these figures, we see clear differences, both between functions and between algorithms. While some algorithms occur in both Figures~\ref{fig:I1_values_subset} and~\ref{fig:I2_values_subset}, many are included only once, indicating that they are relatively good choices for one part of the search, but not the remainder. The clearest example of this is HMLSL~\cite{HMLSL2013}, which performs very well as $A_1$, but has relatively high $I_2$-values. This is caused by the fact that this algorithm typically converges quickly to a value close to the optimum, but has issues in the final exploitation phase, thus only being beneficial to use at the start of the search. We also notice that in general, the $I_2$-values are much lower across all algorithms, indicating that the choice of starting algorithm is the most important for dynAS, while most good algorithms can provide similar benefits to the final part of the search. 

% We can aggregate this data over all functions to get an even more general overview of which algorithms perform well in the initial exploratory stage and which perform well in the final part of the search. This is done by averaging the $I_1$ and $I_2$ values respectively over all functions. An algorithm needs to be admissible on at least 20 functions to be used in this aggregation. %(for a minimum of 20 functions \tb{What does this mean?}).  
% This aggregated data is shown in Figure~\ref{fig:boxplotI1I2_5D}. In this figure, we can see that most algorithms have a lower $I_2$ value, so they give more benefit in the second part of the search. There is however one exception, which is HMLSL~\cite{HMLSL2013}. This algorithm typically converges quickly to a value close to the optimum, but has issues in the final exploitation phase, thus being very beneficial to use at the start of the search.


\subsection{Small Portfolio: Case Study}\label{sec:small_portfolio}

Since the algorithm space we consider is quite large, it can be challenging to gain insights into the individual algorithms. To show that dynamic algorithm selection is also applicable to smaller portfolio's, we limit ourselves to 5 algorithms. These are representative of some widely used algorithm families: Nelder-Doerr~\cite{nelderdoerr}, DE-Auto~\cite{DE_AUTO}, Bipop-aCMA-Step~\cite{bipop_acma_step}, HMLSL~\cite{HMLSL2013} and PSO-BFGS~\cite{PSO_BFGS}.%\died{Do we need references to these algorithms?}\carola{yes, would be good. I think you can find many of them in~\cite{hansen2010comparing,KerschkeT19}}. 
With this reduced algorithm portfolio, we can study the improvements over their respective \VBSs in more detail, and find interesting algorithms combinations to explore further.

\begin{figure}
    \centering
    \includegraphics[width=0.48\textwidth, trim={0, 7, 0, 15},clip]{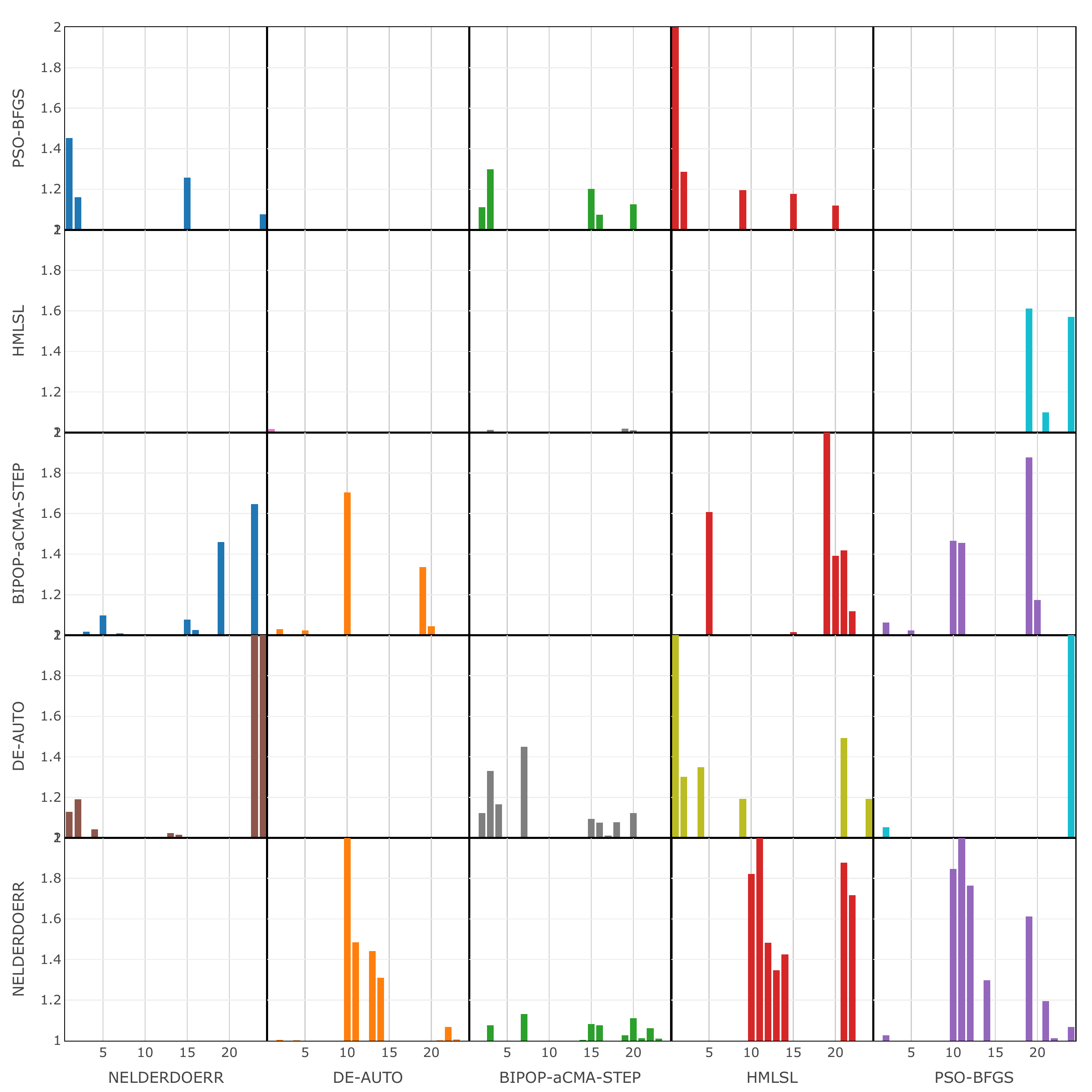}
     \caption{Overview of the best possible ERTs of the combination of algorithms $A_1$ and $A_2$ over \VBSs. Each plot represents a single $A_1$ (X-axis), $A_2$ (Y-axis) combination, where each bar represents a single function, in dimension 3. Values are capped at 2.} %\carola{Values are capped at 2 (?)}}% \carola{How about we drop the meaning of which algo is first and which is second, and only show how much the best combination of the two algos would achieve? This would reduce the size by 2, and if we strivctly want, we can still use a small symbol or color to indicate which one is A1, which one A2. I would probably also change the selection of algos and remove those for which the whole line/column is almost white}}
    \label{fig:matrix_of_bars}\vspace{0pt}
\end{figure}

In Figure~\ref{fig:matrix_of_bars}, we show the relative improvement in ERT over \VBSs of the best combination of two algorithms. In each subplot, all 24 functions are represented. Note that the diagonal represents the static algorithms, which can never lead to an improvement over the \VBSs. We notice some clear trends in this figure. Specifically, we notice that using HMSLS as $A_2$ is rarely effective, while it provides large benefits when used in the initial part of the search. We also note that Nelder-Doerr has the reverse behaviour, seemingly performing much better in the final exploitation phase. 

\begin{figure}
    \centering
    \includegraphics[width=0.5\textwidth, trim={0, 15, 0, 10},clip]{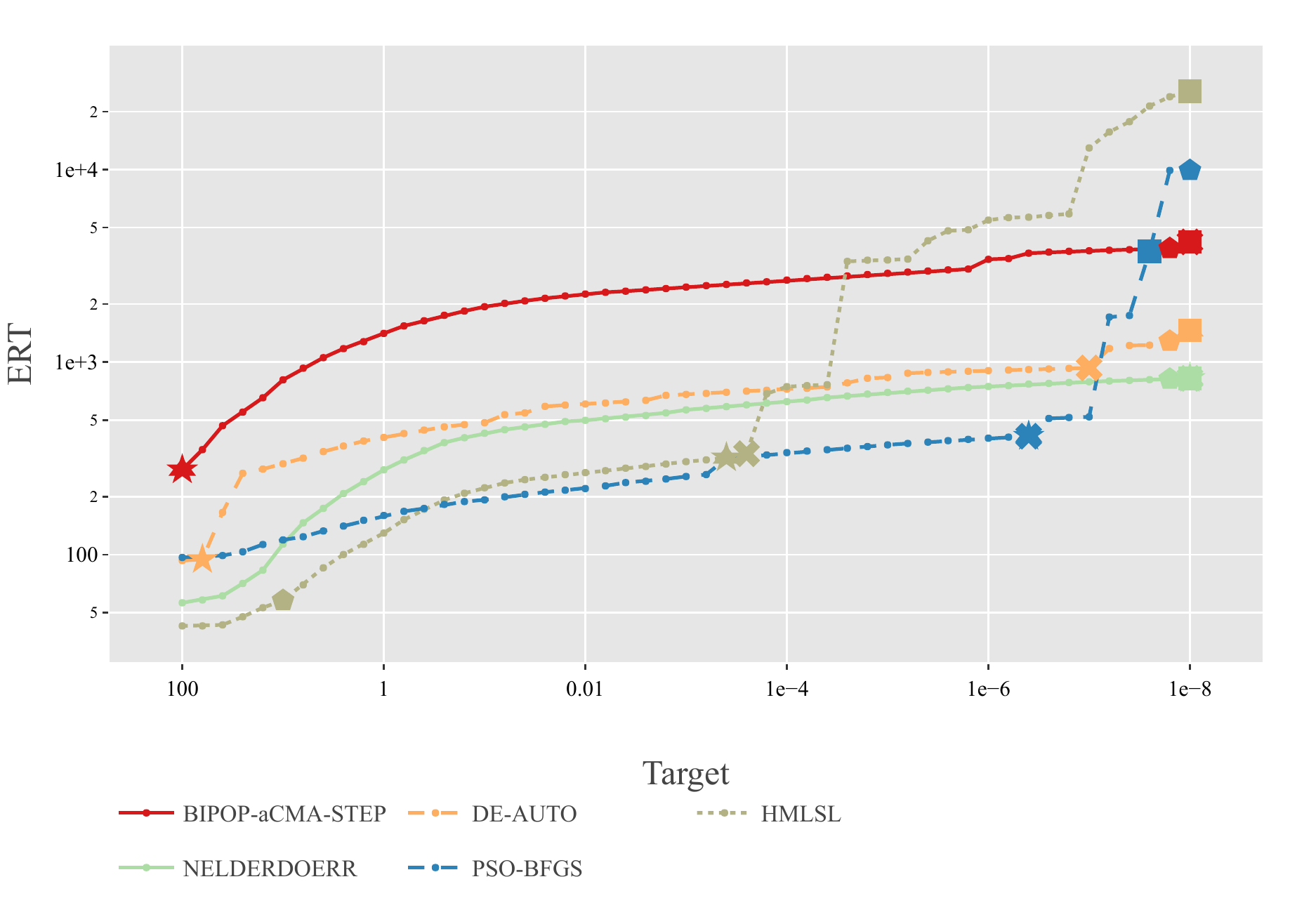}
    \caption{ERT-curves for a selected algorithm portfolio of size 5 on F12 in 3D. Markers indicate optimal switch points between algorithms. Their color and symbol indicate the starting and finishing algorithms respectively.% and the marker symbol represents the algorithm to switch to 
    (star = Nelder-Doerr, triangle = DE-AUTO, cross = BIPOP-aCMA-STEP, square = HMLSL and pentagon = PSO-BFGS).}\vspace{-5pt}
    \label{fig:ert_curve}
\end{figure}

To illustrate the configuration switches which can be considered in this algorithm portfolio, we can zoom in on function 12 in dimension 3 and look at the fixed-target curve showing ERT. This is done in Figure~\ref{fig:ert_curve}, where we also indicate the best switching points between algorithms. This figure highlights the different behaviors of the algorithms in the portfolio, and thus indicates where switching algorithms would be beneficial. The best possible switch in this function would occur from PSO-BFGS to Nelder-Doerr, at target $10^{-6.4}$, leading to a relative speedup of $1.76$ over \VBSs.
%\died{TODO: elaborate on figure once we agree on the portfolio}

To decide which algorithms to use in an algorithm portfolio such as the one used here, two main ways of selecting the algorithms are possible. The first is to use some knowledge about the algorithms to determine which are important. This is useful for initial exploration, but might lead to useful algorithms being ignored. Instead, one can use performance information, such as the $I_1$ and $I_2$-values, to provide some initial representation of the usefulness of algorithms to the portfolio. This approach is much more generic, however the choice of measures can be challenging. For example, the $I_1$ and $I_2$ measures are hard to extend to more general $k$-switch dynAS methods. Instead, an extension of marginal contributions~\cite{XuHHL12} and related concepts such as measures building on Shapley values (like those suggested in~\cite{FrechetteKMRHL16}) would capture algorithm contribution to a portfolio in a much more robust sense, and thus be useful additions to the dynAS setting. %\died{Is this the right location to put this paragraph, or should it be before this section instead?}

\section{Discussion and Future Work}
\subsubsection*{Summary}
The previous results have shown that there is still a large amount of improvement possible over the \VBSs by using dynamic algorithm selection. We have shown several methods to gain insights into the differences between different algorithms and functions. However, the results shown in the previous sections rely on an underlying assumption of feasibility of algorithm switching. For many algorithms, this switching mechanism can be implemented in a relatively straightforward manner, i.e. between different population-based algorithms, such as different CMA-ES variants, for which the algorithm switching methods have already been implemented~\cite{research_project}. 

\subsubsection*{Warm-start} For other algorithms combinations, a dynamic switch during the optimization procedure might be more challenging. For example, a switch from a single-solution algorithm to a population-based one gives rise to an information deficit, which needs to be dealt with to properly initialize the new population. Because of this, the gains indicated by simply combining the ERT values might be tough to achieve in practice. 

More generally, internal parameters are different between algorithms. So the first  challenge to overcome is that one needs to decide how to ``warm-start'' the algorithms, to assure an optimal internal state for the required phase of the optimization process. To be able to achieve the performance of the \VBSd, such warm-start techniques will need to be implemented without the need of additional function evaluations, which could be a big challenge. We would considering to use reinforcement learning approaches to be a promising first step for this task, but since those are quite expensive in terms of computational cost, we hope to see other approaches evolve in the near future.

% The big question is how to materialize the gains indicated by simply combining the ERT values, as done for this study. The first challenge to overcome is that one needs to decide how to ``warm-start'' the algorithms, i.e., how to set their internal control parameters such as, for example, the covariance matrix in the CMA-ES. 

\subsubsection*{Stochasticity} Assuming such warm-start mechanisms are implemented, as was previously done for example within CMA-ES, it has been shown that the theoretical improvements can still be tough to achieve in practice~\cite{research_project}. This is largely caused by the fact that hitting times are stochastic with relatively large variances, which can cause ERT to be unstable. When selecting the $(A_1,A_2,\tau)$-triple, differences in ERT might be obscured by the variance of the hitting times, leading to a worse performance than expected. These effects might become even more important when dealing with larger algorithm spaces, or when incorporating hyperparameters in the search (see paragraph \emph{Hyperparameter tuning}). %, as dynAC approaches will ultimately need to do. \carola{
Analyzing the robustness of common solvers therefore seems to be an essential building block for the development of reliable dynAC approaches.
%Because of this, a study of robustness of performance measures might be required before using them in dynAC contexts. 

\subsubsection*{Switch point} Another challenge which needs to be overcome to achieve effective dynamic algorithm selection is the question how to identify suitable switching points. In this work we used target precision, which is usually not applicable in practice, since the algorithm has no knowledge about the precise value of the optimum. Because of this, we would need to find some other way to use the knowledge of the algorithm to determine when to switch, i.e., the state of internal parameters, landscape features computed from additionally or previously evaluated points, the evolution of fitness values, population diversity, etc.

\subsubsection*{True dynamic switching} While improving the way a switching point is detected is a big challenge to overcome, it also provides new opportunities to improve performance. The estimates shown in this paper consider only a single algorithm switch, whereas a truly dynamic approach could benefit from switching more often, to fully exploit the differences in search behaviour of the different algorithms. %on different parts of the optimization process. 
% To fit this purpose, techniques in which the switches are learned or extracted by the algorithm itself or some other component such as local landscape analysis would seem to be quite promising.

% A related point to note is the fact that This leads to two natural extensions. The first of these is moving to a truly dynamic approach, where the switches are learned or extracted by the algorithm itself or some other component such as local landscape analysis. This can have the added benefit of reducing the impact of ERT-based variance, since switching based on precision to the optimal is both unrealistic in a black-box scenario and not unique for different parts of the landscape, i.e. two separate search-behaviours might reach the target at the same number of evaluations, but be in a completely different part of the landscape.

\subsubsection*{Hyperparameter tuning} A second factor of improvement can come from adding hyperparameter tuning into the dynamic process; i.e., when moving from the algorithm selection setting to a dynamic variant of \emph{Combined Algorithm Selection and Hyperparameter optimization} (CASH~\cite{thornton2013autoweka,Vermetten20CASH}). A dynamic CASH approach would allow the algorithms to specialize even more, so they can focus even more on performing as good as possible on their specific part of the optimization process. %The combination of the added hyperparameter tuning and better switching behaviour would turn the single-switch dynamic algorithm selection as discussed in this paper into a general dynamic algorithm configuration process, which has the potential to significantly improve the final performance.

% While have shown that there are still many challenges to overcome to achieve a succesful implementation of dynamic algorithm selection, the potential benefits are significant. Logically, the next step in this line of research would be implementing some basic dynAS techniques to verify the potential gains shown in this work.

% We have shown that there is a lot of potential performance to be gained by implementing dynamic algorithm selection~(dynAS) on the algorithms from the BBOB datasets. Logically, the next step in this line of research would be implementing such dynAS techniques. However, there are still several challenges which need to be overcome before dynAS can be successfully implemented.

\subsubsection*{Extensions} As any benchmark study, our results are -- for the time being -- limited to the 24 noiseless BBOB functions. Extending them to other classes of numerical black-box optimization problems forms another important avenue for future research. In this context, we consider supervised learning approaches building on exploratory landscape analysis~\cite{mersmann2011exploratory} as particularly promising. It has previously been shown to yield  promising results for the task of configuring the hyper-parameters of CMA-ES~\cite{BelkhirDSS17}.
Note, though, that all existing studies concentrate on static algorithm configuration and/or selection. We would therefore need to extend exploratory landscape analysis to the dynamic setting. First steps into this direction have been made in~\cite{jankovic2019adaptive}, where it is shown that the fitness landscapes, as seen by the algorithm, can change quite drastically during the run.

\subsubsection*{Short-term} All the objectives listed above are quite ambitious. We therefore also formulate a few short-term goals for our research. Building on the techniques used to select interesting algorithms in Section~\ref{sec:small_portfolio}, we aim to create smaller algorithm portfolio's of algorithms for intial implementations of dynAS. This could be done based on techniques studied in this paper, or using measures like the Shapley value~\cite{FrechetteKMRHL16}, allowing for much smaller portfolios which nonetheless capture the different performances of the algorithms. With such a portfolio we can then more efficiently carry out research on the problems mentioned above, i.e., how to warm-start the algorithms and how to decide when to switch from one algorithm to another.  

\begin{acks}
% \vspace{1.5ex}
% \textbf{Acknowledgments.} 
 This work has been supported by the Paris Ile-de-France region.
 %, and by a public grant as part of the Investissement d'avenir project, reference ANR-11-LABX-0056-LMH, LabEx LMH.
\end{acks}

\bibliographystyle{ACM-Reference-Format}
\bibliography{references}

\end{document}